\documentclass[10pt,twocolumn,letterpaper]{article}

\usepackage{cvpr}
\usepackage{times}
\usepackage{epsfig}
\usepackage{graphicx}
\usepackage{amsmath}
\usepackage{amssymb}
\usepackage{amssymb}
\usepackage{subfig}

\usepackage[breaklinks=true,bookmarks=false]{hyperref}

\cvprfinalcopy 

\def\cvprPaperID{****} 

\begin{document}

\title{The AAU Multimodal Annotation Toolboxes: \\Annotating Objects in Images and Videos}

\author{Chris H.\ Bahnsen, Andreas M\o gelmose, Thomas B. Moeslund\\
Visual Analysis of People Laboratory, Aalborg University, Denmark\\
{\tt\small cb@create.aau.dk, andreas@moegelmose.com, tbm@create.aau.dk}
}

\maketitle

\begin{abstract}
    This tech report gives an introduction to two annotation toolboxes that enable the creation of pixel and polygon-based masks as well as bounding boxes around objects of interest. Both toolboxes support the annotation of sequential images in the RGB and thermal modalities. Each annotated object is assigned a classification tag, a unique ID, and one or more optional meta data tags. The toolboxes are written in C++ with the OpenCV and Qt libraries and are operated by using the visual interface and the extensive range of keyboard shortcuts. Pre-built binaries are available for Windows and MacOS and the tools can be built from source under Linux as well. So far, tens of thousands of frames have been annotated using the toolboxes.
\end{abstract}

\section{Introduction}
The main driver behind modern computer vision systems is annotated data - and lots of if. If one wants to train, test, benchmark or verify any vision algorithm that addresses a real-world problem, you need real-world annotated data. You might be lucky that a suitable dataset for your problem exists but often you will need new annotated data that suits your domain.
For many years, this has been the case for most of our work at the Visual Analysis of People Laboratory at Aalborg University. Through a collaborative effort at our lab, we have created two separate annotation tools that can be compiled to run under Windows, MacOS, and Linux. 

\begin{figure}[!t]
    \centering
    \subfloat[Bounding box annotation in RGB] {\includegraphics[width=0.45\columnwidth]{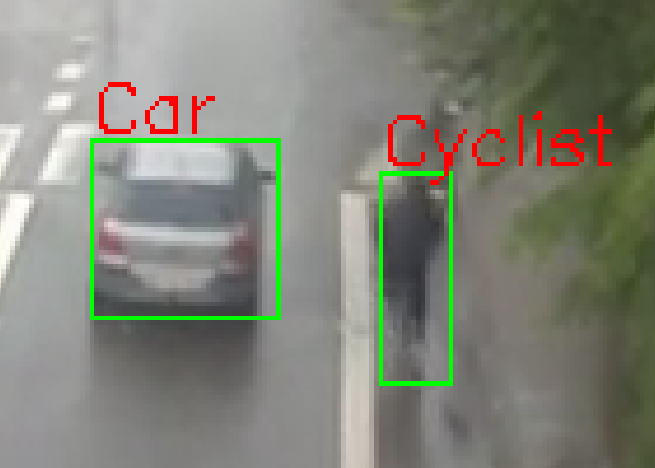}
    \label{fig:bbcar}}
    \hfill
    \subfloat[Corresponding bounding box annotation in thermal]{\includegraphics[width=0.45\columnwidth]{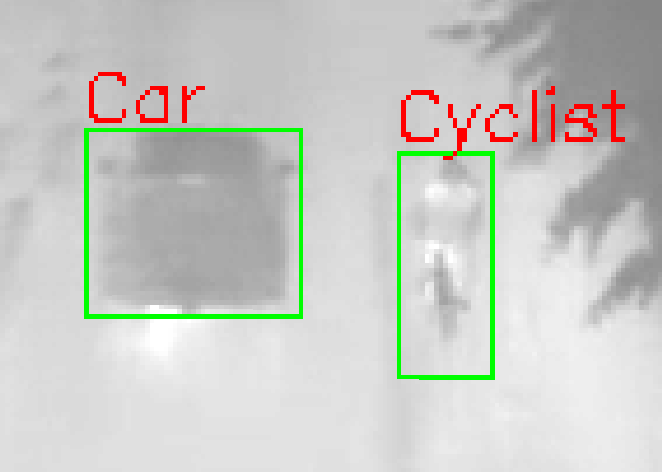}
    \label{fig:thermalBBCar}}
    \hfill
    \subfloat[Pixel annotation in RGB]{\includegraphics[width=0.45\columnwidth]{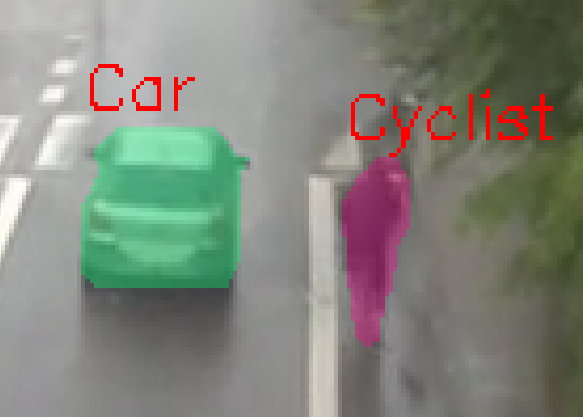}
    \label{fig:pixelcar}}
    \hfill
    \subfloat[Corresponding pixel annotation in thermal]{\includegraphics[width=0.45\columnwidth]{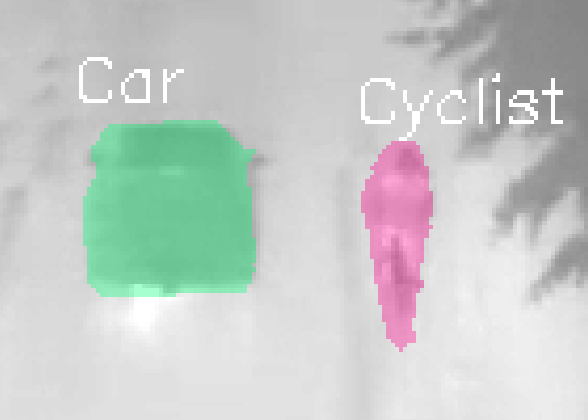}
    \label{fig:thermalPixelCar}}
    \centering
    \caption{Bounding box and pixel-based samples of the same objects annotated in both RGB and thermal modalities. Every annotation is associated with a corresponding tag.}
    \label{fig:annotationsamples}
\end{figure}

The \textit{AAU VAP Multimodal Pixel Annotator} may be used to annotate pixel-based masks of object instances whereas the \textit{AAU VAP Bounding Box Annotator} may be used to annotate bounding boxes around objects of interest. Both annotation tools support annotation tags such that an annotated object may be associated with a predefined class name. Example annotations, both pixel-based and bounding box-based, are shown in Figure \ref{fig:annotationsamples}. 

In this text, we will give an overview of the two annotation tools and the features they provide. An updated list of all annotation tools offered by our laboratory is found at Bitbucket\footnote{ \url{https://bitbucket.org/account/user/aauvap/projects/AN}}. The source code and binaries of the two annotation tools are available under the MIT license. 

The annotation tools have been used to annotate humans \cite{alldieck2017optical, sanguesa2017identifying}, road users \cite{alldieck2016context}, road signs \cite{mogelmose2014traffic}, chicken entrails \cite{philipsen2018organ}, pigs, fish \cite{karpova2018reidentification}, material defects, and more. The number of annotated frames in the examples above vary from a few hundred to tens of thousands.
In the next section, we will describe the common features of the two annotation tools. Section \ref{sec:boundingboxannotator} describes the specific features of the Bounding Box Annotator whereas Section \ref{sec:multimodalpixelannotator} gives a description of the Multimodal Pixel Annotator. Section \ref{sec:conclusion} concludes the work so far and gives insights on the future development of the toolboxes.


%

\section{Common Features}
%
The annotation tools are developed in C++ with Qt and OpenCV \cite{opencv_library} as the main libraries. Both tools have been developed in parallel and thus share many features and much of the code base. The shared features are described below.

\subsection{Object Properties}
\begin{figure}
    \centering
    \includegraphics[width=0.75\columnwidth]{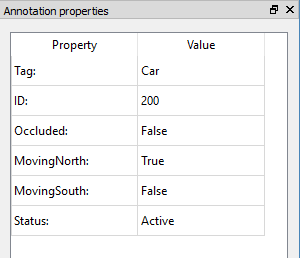}
    \caption{Object properties of an annotation. The "Occluded", "Moving North", and "Moving South" entries are meta data tags that may be either true or false.}
    \label{fig:annotationPropWindow}
\end{figure}

Every annotated object is associated with a unique identification number (ID), a class tag, and optionally one or more meta data tags. An example hereof is shown in Figure \ref{fig:annotationPropWindow}. 

We will go through the object properties below. Properties shown in \textbf{bold} are mandatory whereas properties shown in \textit{italics} are optional.

\begin{itemize}
    \item \textbf{Tag} The class name of an object. The class name may be freely chosen or limited to a pre-defined list if the setting \texttt{Limit annotation tags to suggested list} is checked. The suggested list is populated from the existing annotation tags in the dataset and from the user-editable list available in \texttt{File} $\rightarrow$ \texttt{Edit suggested tags}.
    \item \textbf{ID} The identification number of the object. In Bounding Box Annotator, this number is defined in the range $[0,\inf]$ and is unique for the entire annotation sequence. In Multimodal Pixel Annotator, the ID is encoded into the mask image which limits the range to the interval from $[0,255]$. However, the ID's in the range from [0,10] are reserved for internal operations of the program whereas ID 170 is reserved for don't care borders.
    \item \textit{Meta data tags} The meta data tags are binary object attributes. The meta data names themselves may be specified before creating an annotation sequence in \texttt{File} $\rightarrow$ \texttt{Edit meta data fields} or retrospectively applied by manually editing the csv-file containing the annotations. Three meta data names have been set in Figure \ref{fig:annotationPropWindow}: the "Occluded", "Moving North", and "Moving South" tags. These tags may be either true or false for an object and are defined for every frame. 
    \item \textbf{Status} When annotating video data as described in Section \ref{subsec:annvideodata}, one might choose to copy existing annotations to temporally adjacent frames. However, an object might be moving out of the image frame and as a result, the annotated mask belonging to this object should not be copied to the next frame. This might be changed by setting the object status from \texttt{Active} to \texttt{Last frame reached}.
\end{itemize}

\subsection{Annotation of Sequential Data}
\label{subsec:annvideodata}
The annotation toolboxes assume that the source images are in the same folder. The toolboxes do not directly support video files, mainly because OpenCV does not provide efficient and accurate temporal search for videos. Instead, videos may be converted to a collection of single frames by an FFMPEG script\footnote{\texttt{ffmpeg -i file.mpg -r 1/1 \%05d.png}}. One may configure the annotation toolboxes such that they only load frames that adhere to a specific file pattern. The option is set in  \texttt{Settings} $\rightarrow$ \texttt{File patterns} and supports regular expressions. For simple use cases such as including all .png-files, the string \texttt{*.png} is sufficient.

\begin{figure}
    \centering
    \frame{\includegraphics{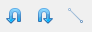}}
    \caption{Buttons from left to right: (1) Retain image when loading previous frame, (2) Retain image when loading next frame, (3) Interpolate between annotations when stepping $>1$ frames.}
    \label{fig:retainimagesbutton}
\end{figure}

\paragraph{Retaining annotations in adjacent frames} When annotating frames that are temporally consistent, i.e.\ the same objects are moving slowly from frame to frame, it might be useful to copy the annotations from frame $n$ to frame $n+1$ or $n-1$. This functionality is found in the \texttt{Retain when loading previous} and \texttt{Retain when loading next} buttons illustrated in Figure \ref{fig:retainimagesbutton}. 

\subsection{Multi-Modal Annotation}
Both annotation tools support the annotation of objects in two views and given the preference in our lab for multi-modal approaches \cite{mogelmose2013tri}, we refer to view 1 as RGB and view 2 as thermal. The RGB modality is the master modality and all annotations are by default stored in a coordinate system relative to the RGB image coordinates. For compatibility with the AAU Trimodal People Segmentation Dataset\footnote{\url{https://www.kaggle.com/aalborguniversity/trimodal-people-segmentation}}, the Multimodal Pixel Annotator also enables a depth modality which is currently in legacy support. 

Registration from $\text{RGB} \leftrightarrow \text{Thermal}$ can be performed using a single homography which may be sufficient if the objects of interest in the scene are lying in close proximity to the same plane. The homographies should be stored in a yml-file using the OpenCV FileStorage method in the \texttt{homRgbToT} and \texttt{homTToRgb} variables. Example homographies are found from the sample annotations provided at the Bitbucket project pages. 

If the planar constraint is violated and a single homography is not sufficiently accurate, one may use a combination of multiple homographies. More details about this approach are found in the work by Palmero et al.\ \cite{palmero2016multi}. 

\subsection{Don't Care Masks}
\begin{figure}
    \centering
    \includegraphics[width=0.85\columnwidth]{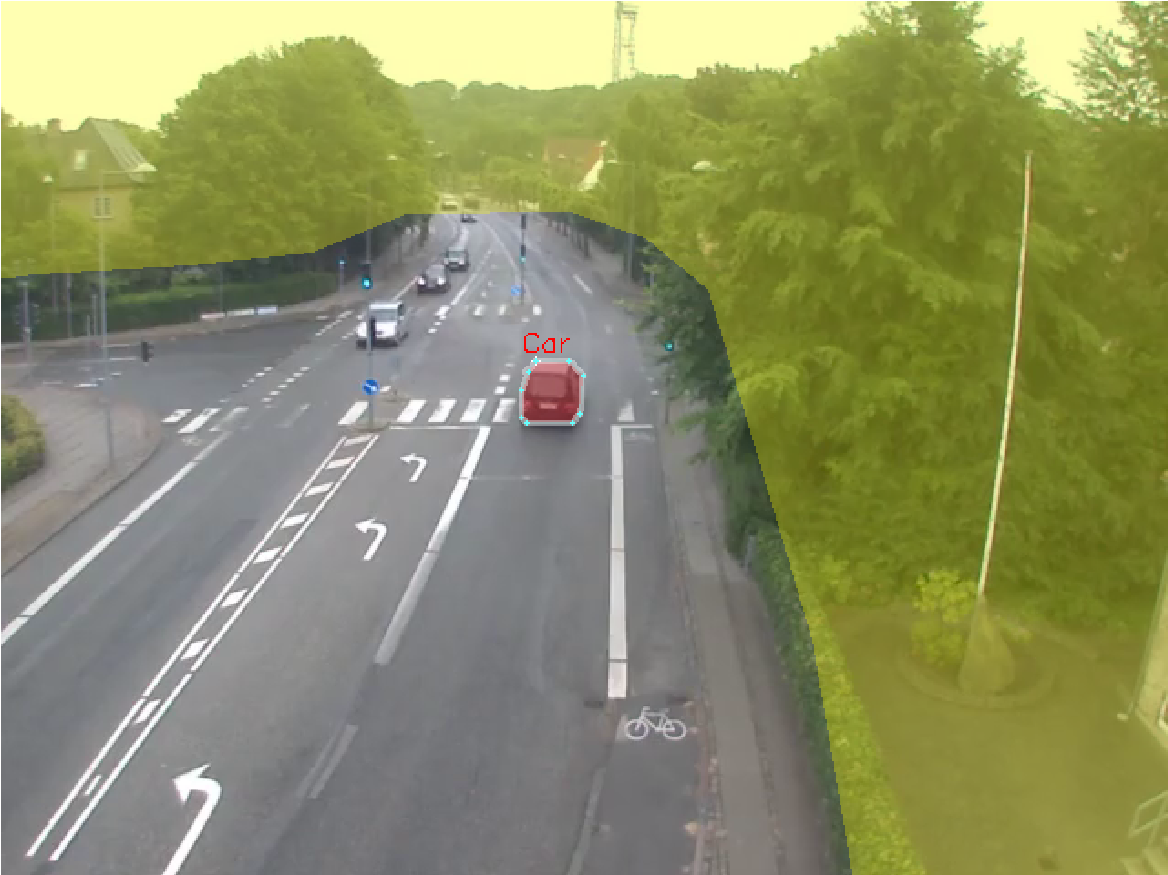}
    \caption{The don't care mask of the image is overlaid in yellow. The colour and opacity of the mask may be defined by the user.}
    \label{fig:dontCareMask}
\end{figure}

It might be beneficial to use a don't care mask that visualizes the region-of-interest in which objects should be annotated. If this option is enabled in settings, a binary mask image should be placed in the root folder of the annotations or the directory above. If the don't care mask is placed here under the name \texttt{mask.png}, the mask will be loaded automatically when opening an annotated sequence. An example of a don't care mask is shown in Figure \ref{fig:dontCareMask}.

\subsection{Shortcut-driven Annotations}
Maximizing the use of the keyboard is one of the better ways of speeding up the annotation process. Besides the mouse-driven drawing functionality, almost every other aspect of the annotation tools may be operated by using the keyboard. The respective shortcuts are revealed by hovering the mouse on top of each button. Alternatively, the wiki pages\footnote{\url{https://bitbucket.org/aauvap/bounding-box-annotator/wiki/Home}}$^{,}$\footnote{\url{https://bitbucket.org/aauvap/multimodal-pixel-annotator/wiki/Home}} of the annotation tools provide a great overview of the available shortcuts.

\section{Bounding Box Annotator}
\label{sec:boundingboxannotator}
The Bounding Box Annotator provides an interface for drawing bounding boxes around objects of interest. It provides additional features for working with image sequences such as interpolation and extended annotation deletion and merging functionality.

\subsection{Temporal Interpolation}
\label{subsec:temporalinterpolation}
\begin{figure*}
    \centering
    \includegraphics[width=0.99\textwidth]{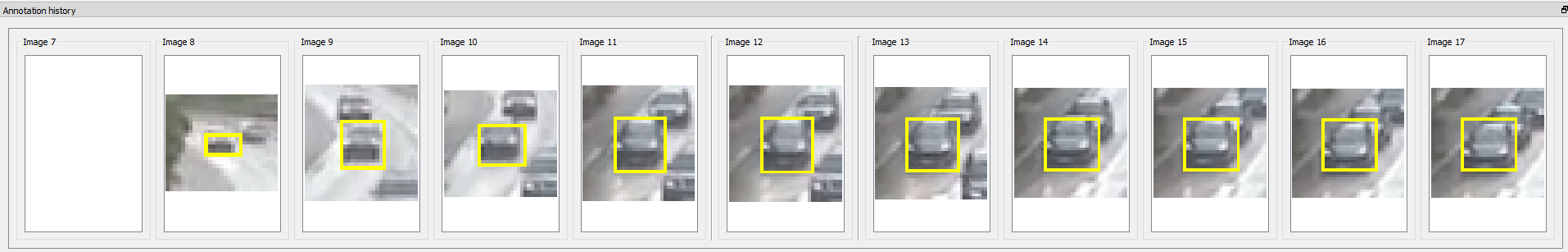}
    \caption{The annotation history window of the Bounding Box Annotator. The selected annotation of the current frame (Image 12) is shown in the middle, surrounded by annotations containing the same ID in the previous and next five frames. Image 7 is empty, indicating that the object ID does not exist in this frame.}
    \label{fig:annotationhistorywindow}
\end{figure*}

When working with image sequences with high frame-rate and slow-moving objects, annotating every single frame is usually a very tedious task. The Bounding Box Annotator attempts to ease the annotation process by:
\begin{itemize}
    \item Providing an overview of annotations with the same ID in the neighbouring frames, illustrated in Figure \ref{fig:annotationhistorywindow}.
    \item Interpolating between annotations. If the user annotates an object in frame 1 and frame 6, the program optionally interpolates between these annotations to create corresponding annotations for frame 2, 3, 4, and 5. Best results are achieved when the motion of the object is nearly linear.
\end{itemize}

\subsection{Deleting and Merging Annotations}
\begin{figure}
    \centering
    \frame{\includegraphics[width=0.70\columnwidth]{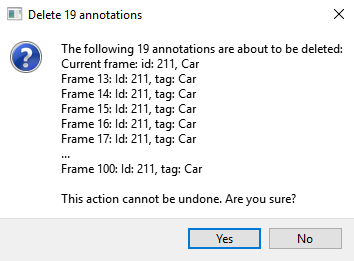}}
    \caption{Deleting annotations with ID 211 in the current and subsequent frames. The user is asked to acknowledge the severity of this action before deletion.}
    \label{fig:deleteAnnotationsPrompt}
\end{figure}

When using the 'retain image' buttons illustrated in Figure \ref{fig:retainimagesbutton}, one might forget to set the \texttt{Last frame reached} flag, leading to several duplicate annotations once the object of interest has left the frame. The button \texttt{Delete selected annotations in current and future frames} comes to the rescue, effectively deleting annotations with the selected ID(s) in all future annotations. The program will inform the user about the affected annotations, hopefully minimizing the risk of deleting a bunch of annotations by accident. A sample prompt is shown in Figure \ref{fig:deleteAnnotationsPrompt}.

\begin{figure}
    \centering
    \frame{\includegraphics[width=0.99\columnwidth]{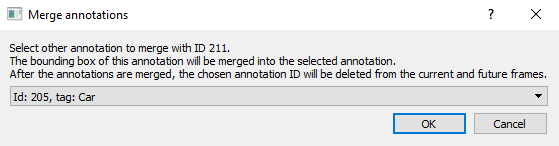}}
    \caption{Merging an annotation ID with the currently selected annotation ID in the current and subsequent frames.}
    \label{fig:mergeAnnotationsPopUp}
\end{figure}

Two annotations might be merged by using the \texttt{Merge selected annotation and another annotation in current and future frames} button, which will do just that. After merging, the original 'other' annotation will be deleted as described in Figure \ref{fig:mergeAnnotationsPopUp}.

\subsection{Automatic Backup}
The .csv-file containing the annotations is automatically copied to a backup folder whenever an annotation folder is opened with the Bounding Box Annotator. The backup file is timestamped such that the user may easily revert to an older revision if the current annotations are deleted by accident.

\subsection{Exporting Annotations}
The Bounding Box Annotator saves the annotations in a single file, by default named \texttt{annotations.csv}. Each annotated object represents a line in the csv-file and the bounding box is encoded by saving the pixel coordinates of the upper left corner and the lower right corner. However, it is unlikely that this is the format of your favourite machine learning algorithm.

Currently, the Bounding Box Annotator is capable of exporting the annotations to the format used by the YOLO network running on Darknet \cite{redmon2017yolo9000}. 
When training a network on Darknet, every image should have a corresponding annotation file where each line indicates the category ID, centre point (X,Y), width, and height of an annotated object, all in normalized image coordinates\footnote{Curiously, the output format of YOLO/Darknet is not the same as the input format.}. The tag of an annotated object is translated to the corresponding category ID by selecting an appropriate category list. Out of the box, the tool comes with category lists for MSCOCO \cite{lin2014microsoft}, ImageNet-1000 \cite{deng2009imagenet}, YOLO-9000 \cite{redmon2017yolo9000}, and PASCAL VOC \cite{everingham2010pascal}. If one wants to use his own list, it can be added in the \texttt{categoryLists} folder in the root directory of the program. 

\section{Multimodal Pixel Annotator}
\label{sec:multimodalpixelannotator}
The Multimodal Pixel Annotator allows fine-grained pixel-level annotations. The specific functionality of the annotation tools is described below.

\begin{figure}
    \centering
    \frame{\includegraphics[width=0.99\columnwidth]{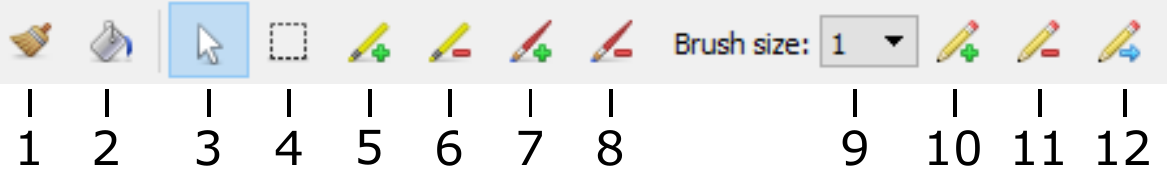}}
    \caption{Drawing tools in Multimodal Pixel Annotator. The numbers refer to the following: 
    \\1) Removing noise from the mask.
    \\2) Filling holes in the mask.
    \\3) Selecting an annotation.
    \\4) Initializing GrabCut.
    \\5-6) Adding true positive/negative brushes to the GrabCut mask.
    \\7-8) Manually add to/remove from mask.
    \\9) Define brush size of tools 5-8.
    \\10-12) Add/remove/move point from polygon mask.}
    \label{fig:pixelLevelTools}
\end{figure}

\subsection{Drawing the mask}
The user has three options for drawing a mask using the pixel annotation tool: 
\begin{enumerate}
    \item Initializing the mask and refining it using GrabCut \cite{rother2004grabcut}.
    \item Using paint-style brush tools.
    \item Defining a contour around the object of interest using the polygon tool.
\end{enumerate}

The graphical buttons for drawing the mask are shown in Figure \ref{fig:pixelLevelTools}.

\subsubsection{Using GrabCut}

\begin{figure}[!t]
    \centering
   \captionsetup[subfigure]{justification=centering}
    \subfloat[Initializing GrabCut]{\includegraphics[width=0.31\columnwidth]{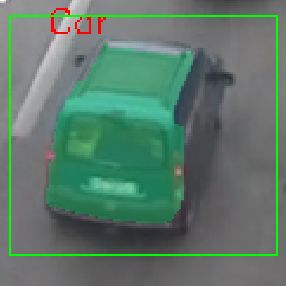}
    \label{fig:grabCutInitialization}}
    \subfloat[Adding true positives (red)]{\includegraphics[width=0.31\columnwidth]{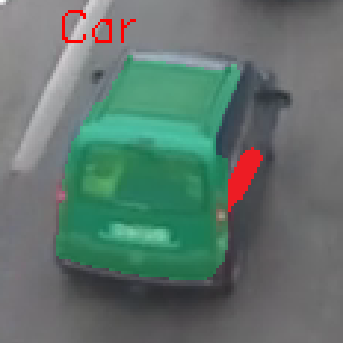}
    \label{fig:grabCutAddToMask}}
    \subfloat[The resulting GrabCut mask]{\includegraphics[width=0.31\columnwidth]{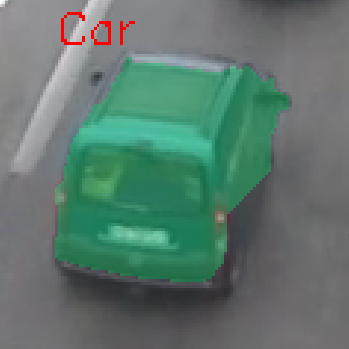}
    \label{fig:grabCutResult}}
    \caption{Example use of the GrabCut tools. Steps b)-c) are performed iteratively until the mask covers the object of interest.}
    \label{fig:grabCutIllustration}
\end{figure}

When using GrabCut, the user should initialize a bounding box around the object of interest. If the appearance of the object is significantly different from the background, chance is that the initial GrabCut segmentation may be good enough. If that is not the case, the user may supply ground truth positive and negative brushes to guide the GrabCut segmentation. An example is shown in Figure \ref{fig:grabCutIllustration}. Please keep in mind that GrabCut segmentation is an iterative process and the entire mask may change whenever true positive and negative brushes are drawn. If one wants to apply final touches to an otherwise finished mask, the manual brush tools should be used. 

\subsubsection{Manually Painting the Mask}
If the segmentation results of the GrabCut approach are not satisfactory, the manual brush tools may be used instead. A variety of different brush sizes are provided to fit the size of the object of interest.

\subsubsection{Drawing Polygons}
\begin{figure}
    \centering
    \includegraphics[width=0.85\columnwidth]{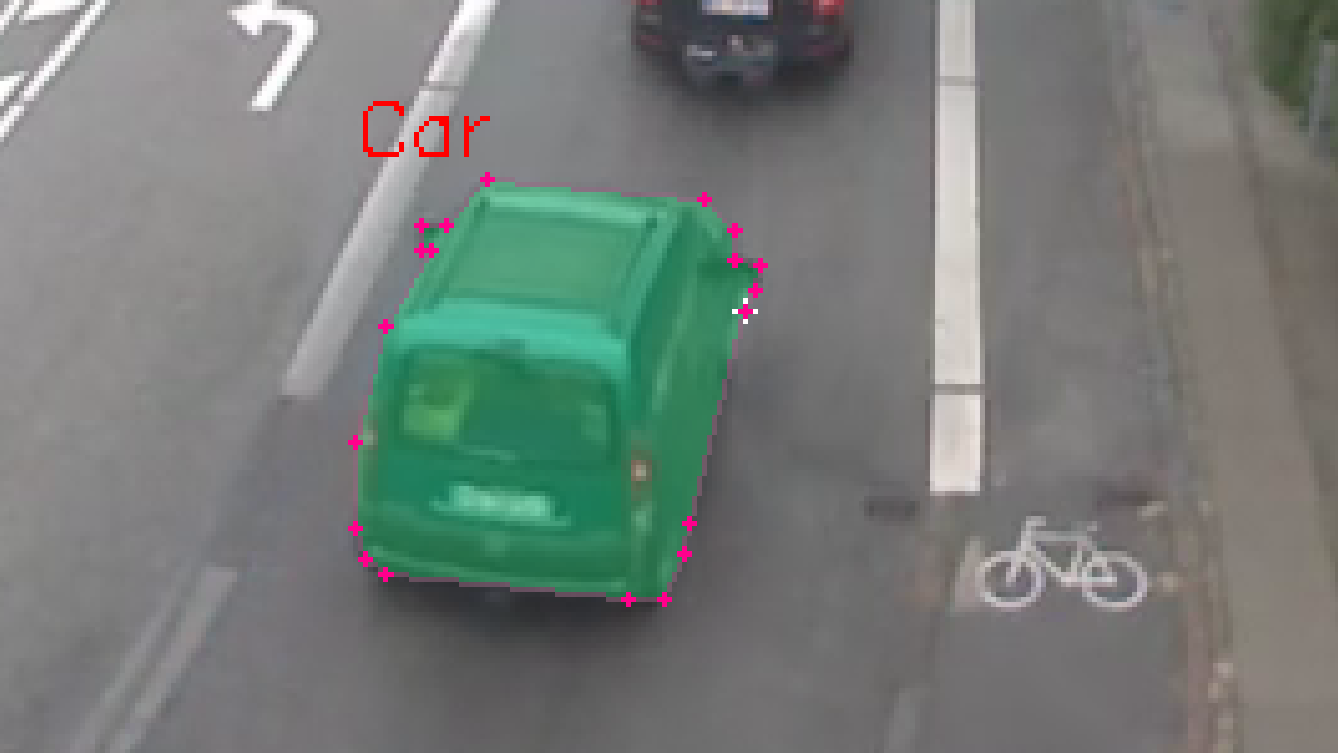}
    \caption{Drawing a polygon around the annotated object.}
    \label{fig:polygonCarMask}
\end{figure}
If the objects to be annotated are rigid, with well-defined borders and without holes, it might be beneficial to draw the points defining the outer contour of the object. This is made possible by using the polygon tools and placing points around the outline of the object. A sample annotation using the polygon-based tools is shown in Figure \ref{fig:polygonCarMask}.

\subsubsection{Don't Care Borders}
To allow for ambiguous segmentation results around the border of objects, one can add a \textit{don't care border} around the object masks. This option is available as "annotation borders" in \texttt{File} $\rightarrow$ \texttt{Settings} $\rightarrow$ \texttt{Annotations}. The width of the don't care border is also configurable from these settings. The don't care border is encoded in the masks with grey-scale value 170.

\subsubsection{Filtering the Mask}
The annotated mask might contain unwanted noise in the form of isolated pixels or
small holes in the mask. These two problems are often encountered when using the GrabCut tools and can be easily resolved using the \texttt{Remove noise} and \texttt{Fill holes} functions depicted in Figure \ref{fig:pixelLevelTools}.

\subsection{Exporting Annotations}
The Multimodal Pixel Annotator maintains a list of the annotations in a single csv-file, with every annotated object containing one line in the annotation. If only the polygon tools are used, the file is self-contained. On the other hand, annotated masks created using the GrabCut or brush tools are saved as grey-scale images where the annotation ID determines the shade of grey of the mask. In this case, the csv-file keeps track of the image files, the tag names, and the meta data tags.

There are currently two options for exporting the annotations:
\begin{itemize}
    \item Converting the annotations in a bounding box-format supported by the Bounding Box Annotator. 
    \item Exporting the annotations to a format compatible with the COCO API \cite{lin2014microsoft}. This creates a single json-file containing a list of all annotated images, a list of object classes, and a list of annotations either represented as polygons or compressed using run-length encoding.
\end{itemize}

\section{Conclusion and Future work}
\label{sec:conclusion}
This concludes the brief tour of our image annotation tools. The tools have been valuable for many different purposes in our laboratory and we sincerely hope that they will be useful for future annotation projects as well. 
Our laboratory have annotated tens of thousands of frames using the annotation tools and it is our experience that once one gets acquainted with the work-flow and the shortcuts, these tools provide a good environment for hours, weeks, and months of annotation work.
Since the annotation tools are developed as side-line projects during our PhD's, there might be some occasional rough edges when using the programs. If the reader encounters any unexpected behaviour during the use of the programs, he or she is more than welcome to open an issue on Bitbucket.

In the future, we expect to merge the code base of the two annotation programs such that a bounding box annotation is a special case of a polygon-based annotation which again is a special case of a pixel-based annotation. 
If resources and time allow, we might even investigate semi-supervised annotation methods that could speed up the annotation process.

\section*{Acknowledgements}
We greatly appreciate the work of our student annotators during the years and the many hours that they have spent using the programs. Their continued work has uncovered numerous bugs which is critical in developing annotation tools that work as intended.

{\small
\bibliographystyle{ieee}
\bibliography{annotationToolboxes.bib}

\begin{thebibliography}{10}\itemsep=-1pt

\bibitem{alldieck2016context}
T.~Alldieck, C.~H. Bahnsen, and T.~B. Moeslund.
\newblock Context-aware fusion of rgb and thermal imagery for traffic
  monitoring.
\newblock {\em Sensors}, 16(11):1947, 2016.

\bibitem{alldieck2017optical}
T.~Alldieck, M.~Kassubeck, B.~Wandt, B.~Rosenhahn, and M.~Magnor.
\newblock Optical flow-based 3d human motion estimation from monocular video.
\newblock In {\em German Conference on Pattern Recognition}, pages 347--360.
  Springer, 2017.

\bibitem{opencv_library}
G.~Bradski.
\newblock {The OpenCV Library}.
\newblock {\em Dr. Dobb's Journal of Software Tools}, 2000.

\bibitem{deng2009imagenet}
J.~Deng, W.~Dong, R.~Socher, L.-J. Li, K.~Li, and L.~Fei-Fei.
\newblock Imagenet: A large-scale hierarchical image database.
\newblock In {\em Computer Vision and Pattern Recognition, 2009. CVPR 2009.
  IEEE Conference on}, pages 248--255. Ieee, 2009.

\bibitem{everingham2010pascal}
M.~Everingham, L.~Van~Gool, C.~K. Williams, J.~Winn, and A.~Zisserman.
\newblock The pascal visual object classes (voc) challenge.
\newblock {\em International journal of computer vision}, 88(2):303--338, 2010.

\bibitem{karpova2018reidentification}
A.~Karpova and J.~B. Haurum.
\newblock Re-identification of zebrafish using metric learning.
\newblock {\em Unpublished Master Thesis, Aalborg University, Aalborg,
  Denmark}, 2018.

\bibitem{lin2014microsoft}
T.-Y. Lin, M.~Maire, S.~Belongie, J.~Hays, P.~Perona, D.~Ramanan,
  P.~Doll{\'a}r, and C.~L. Zitnick.
\newblock Microsoft coco: Common objects in context.
\newblock In {\em European conference on computer vision}, pages 740--755.
  Springer, 2014.

\bibitem{mogelmose2013tri}
A.~Mogelmose, C.~Bahnsen, T.~Moeslund, A.~Clapes, and S.~Escalera.
\newblock Tri-modal person re-identification with rgb, depth and thermal
  features.
\newblock In {\em Proceedings of the IEEE Conference on Computer Vision and
  Pattern Recognition Workshops}, pages 301--307, 2013.

\bibitem{mogelmose2014traffic}
A.~M{\o}gelmose, D.~Liu, and M.~M. Trivedi.
\newblock Traffic sign detection for us roads: Remaining challenges and a case
  for tracking.
\newblock In {\em Intelligent Transportation Systems (ITSC), 2014 IEEE 17th
  International Conference on}, pages 1394--1399. IEEE, 2014.

\bibitem{palmero2016multi}
C.~Palmero, A.~Clap{\'e}s, C.~Bahnsen, A.~M{\o}gelmose, T.~B. Moeslund, and
  S.~Escalera.
\newblock Multi-modal rgb--depth--thermal human body segmentation.
\newblock {\em International Journal of Computer Vision}, 118(2):217--239,
  2016.

\bibitem{philipsen2018organ}
M.~P. Philipsen, J.~V. Dueholm, A.~J{\o}rgensen, S.~Escalera, and T.~B.
  Moeslund.
\newblock Organ segmentation in poultry viscera using rgb-d.
\newblock {\em Sensors}, 18(1):117, 2018.

\bibitem{redmon2017yolo9000}
J.~Redmon and A.~Farhadi.
\newblock Yolo9000: better, faster, stronger.
\newblock {\em arXiv preprint}, 2017.

\bibitem{rother2004grabcut}
C.~Rother, V.~Kolmogorov, and A.~Blake.
\newblock Grabcut: Interactive foreground extraction using iterated graph cuts.
\newblock In {\em ACM transactions on graphics (TOG)}, volume~23, pages
  309--314. ACM, 2004.

\bibitem{sanguesa2017identifying}
A.~A. Sang{\"u}esa, T.~B. Moeslund, C.~H. Bahnsen, and R.~B. Iglesias.
\newblock Identifying basketball plays from sensor data; towards a low-cost
  automatic extraction of advanced statistics.
\newblock In {\em Data Mining Workshops (ICDMW), 2017 IEEE International
  Conference on}, pages 894--901. IEEE, 2017.

\end{thebibliography}
}

\end{document}